\documentclass[runningheads]{llncs}

\usepackage{makeidx}
\usepackage{graphicx}
\usepackage{amsmath,amssymb} 
\usepackage{color}
\usepackage{epsfig}
\usepackage{graphicx}
\usepackage{cuted}
\usepackage{float}
\usepackage{authblk}
\usepackage{xspace} 
\usepackage{hyperref}
\DeclareUnicodeCharacter{FB01}{fi}
\newcommand{\etal}{\textit{et al.}}
\newcommand{\ie}{i.e.\@\xspace}
\newcommand{\R}{\mathbb{R}}
\newcommand{\N}{\mathbb{N}}

\DeclareMathOperator*{\argmax}{\mathrm{argmax}}

\begin{document}
	\pagestyle{headings}
	\mainmatter

	\title{Towards Better Adversarial Synthesis of Human Images from Text}

	\titlerunning{Towards Better Adversarial Synthesis of Human Images from Text}
	\authorrunning{Briq,Kochar,Gall}
	\author{Rania Briq\textsuperscript{1}
	\and
	Pratika Kochar\textsuperscript{}
	\and
	Juergen Gall\textsuperscript{1}}
	\institute{ Computer Vision Group, University of Bonn\\
	{\tt\small briq,gall@iai.uni-bonn.de}
}

	\maketitle
	
	\begin{abstract}
This paper proposes an approach that generates multiple 3D human meshes from text. The human shapes are represented by 3D meshes based on the SMPL model. The model's performance is evaluated on the COCO dataset, which contains challenging human shapes and intricate interactions between individuals. The model is able to capture the dynamics of the scene and the interactions between individuals based on text. We further show how using such a shape as input to image synthesis frameworks helps to constrain the network to synthesize humans with realistic human shapes.
	\end{abstract}
	
	\section{Introduction}
	\label{sec:introduction}
	\begin{figure*}[t]
		\begin{center}
			\includegraphics[width=12cm]{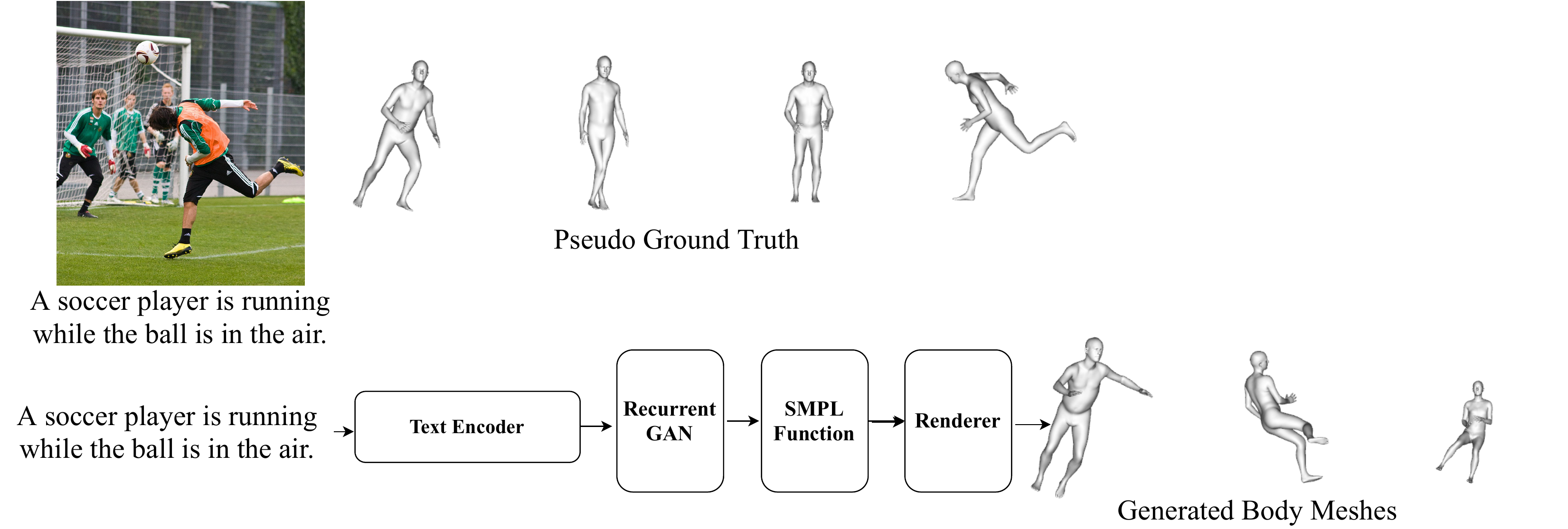}
		\end{center}
		\caption{An example from the COCO dataset with text annotations describing the scene. The pseudo ground truth refers to the SMPL body meshes that are estimated using some pretrained model such as SPIN \cite{kolotouros2019learning}. The proposed model is illustrated in the bottom row, where a text encoder takes a text description and encodes the words to obtain their embeddings. A recurrent GAN is used which takes these embeddings and generates a set of SMPL parameters where each element represent one person shape. The SMPL function computes 3D vertices representing the body mesh and the renderer projects these vertices onto the 2D plane.}
		\label{fig:approach}
	\end{figure*}
\begin{figure}[t]
	\begin{center}
		\includegraphics[width=1\linewidth]{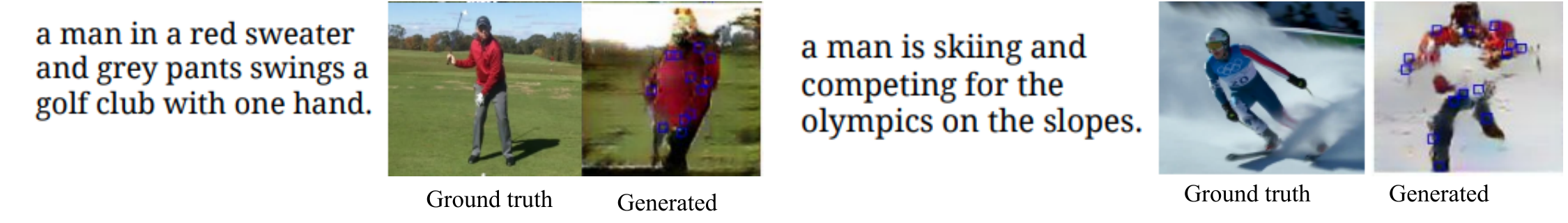}
	\end{center}
	\caption{The input to this GAN from \cite{NIPS2016_a8f15eda} is the 2D keypoints in addition to the text. As can be seen, the generated person is incomplete and distorted.}
	\label{fig:where_to_draw_eg}
\end{figure}

Imagining the overall shape of persons based on a text description describing a scene is an easy task for humans even when it involves multiple humans interacting with each other such as in a football match. However, this task is not as simple for machines since the human shape is highly articulated and is usually in interaction with other individuals and objects in the environment, which creates a huge space of versatility. In this work, we propose a model that can generate 3D human shapes matching a high-level text description. Motivations for creating the 3D human mesh include animating multiple 3D avatars based on text and guiding image-synthesis frameworks towards synthesizing more realistic human shapes. In a similar work, Zhang \etal \cite{zhang2021adversarial} proposes a convolutional neural network (CNN) for generating a single person pose from text, which generates only the 2D joints but not the full shape and 3D pose. The model also has a limited capacity such that it can only generate one pose for any given text, but cannot  handle scenarios that contain multiple poses. A 2D pose may be useful for guiding a pose transfer task as in~\cite{zhou2019text,ma2017pose,tang2020xinggan,tang2020bipartite} where an image of the person is given, but is far too sparse to constrain a network to generate a realistic-looking unseen person from text. Additionally, synthesizing multi-person shapes adds much more complexity to the task, since in typical scenarios the persons in the scene occlude or overlap with each other.
In this work, we address the problem of generating a variable number of 3D human shapes based on the SMPL model \cite{loper2015smpl}. Figure~\ref{fig:approach} describes the overall task. In our scenario, the model should have the capability to deal with a variable number of shapes that match the text description, as well as infer the interactions between the different individuals in the scene. Since a non-recurrent GAN cannot handle content generation with a variable number of outputs, we propose a recurrent generative adversarial network (GAN). Our proposed recurrent GAN consists of a recurrent generator and recurrent discriminator (critic) such that the generator generates a single 3D body shape at each of its iterations, and the critic iterates over a set of shapes and outputs a final score as a many-to-one recurrent network. To help the generator focus on the relevant words in the current iteration, we additionally learn words weights that indicate how much information each word is contributing to generating current shape. The hidden state of the recurrent component maintains information that helps it remember for which words semantically consistent shapes have already been generated and for words more shapes have yet to be generated. The critic has to fulfill two roles, first it has to assess the quality of the individual shapes and their semantic consistency with the given text. Additionally it has to assess the matching between the generated shape and the text in a collective way, so that for example it can tell if certain words are still not satisfied, \ie not all words have a corresponding human shape. 
When a text describes multiple people in the scene or even a single person, image generation methods synthesize images with good-looking scenes but with a disfigured human shape~\cite{xu2018attngan,li2019object,tao2020df,zhu2019dm}. Using 2D keypoints helps slightly as shown in \cite{NIPS2016_a8f15eda} (figure~\ref{fig:where_to_draw_eg}), however, it is too sparse to help guide these frameworks in generating a realistic-looking human.  
To demonstrate the benefit of the proposed task, we integrate our model with an image synthesis framework while using COCO~\cite{lin2014microsoft} as the training dataset and show how it substantially improves the person's appearance in the synthesized images. 
	
	\section{Related work}
	Generating high-fidelity visual content is a crucial task in computer vision. Over the past few years, deep generative models such as \cite{hinton2009deep,salakhutdinov2009deep,kingma2013auto,goodfellow14,salakhutdinov2015learning} have improved the fidelity of generated content to a great extent. Generative Adversarial Networks  (GANs) which were first introduced in~\cite{goodfellow14} have be been a powerful paradigm for learning high dimensional data distributions. In the field of computer vision, GANs have been employed for different tasks for content synthesis, including unconditional image synthesis~\cite{goodfellow14,radford16}, image synthesis conditioned on text~\cite{li2019object,Li_2019_story,NIPS2019_8375,qiao2019mirrorgan,reed16,xu2018attngan,zhang17,zhu2019dm,tan2019semantics,tao2020df,ramesh2021zero}, generating text description conditioned on images~\cite{dai2017towards}, style transfer between images~\cite{chu2017cyclegan},pose transfer in person images ~\cite{ma17,tang2020bipartite,tang2020xinggan,zhou2019text}, and human pose synthesis conditioned on text \cite{zhang2021adversarial}. In \cite{NIPS2016_6111}, the authors generate images based on text and show that using a sparse set of keypoints makes synthesizing a higher resolution image possible. Zhou \etal~\cite{zhou2019text} alter the pose of a person in a given image based on a text description. The approach assumes that all poses can be represented by a set of clusters and is applied to a pedestrian dataset in which the poses are of standing or walking persons, and therefore cannot be applied to datasets with complex poses such as COCO.
	Li \etal~\cite{li2019object} propose an object-driven attention module that relies on an object classifier loss. However, the method fails to generate plausible human shapes despite improved results. 
	In fashion applications, Zhu \etal~\cite{Zhu_2017_ICCV} manipulate the clothing of a person in a given image based on a text description without altering the pose. 
Prokudin \etal \cite{prokudin2021smplpix} propose generating photorealistic images of humans using a sparse set of 3D vertices and a map of their corresponding RGB colors and depth values.
	Other related works such as \cite{grabner2011makes,gupta20113d,Li_2019_affordance,zhang2020generating} deal with searching for or synthesizing plausible human poses that match object affordances in a given scene.
	Recurrent GAN architectures for generating sequential data was introduced in \cite{mogren2016c} and applied to music data. In a subsequent work, Hyland \etal \cite{esteban2017real} included a conditional input and applied it for generating time-series data in the medical domain.

	\section{Approach} 
	\begin{figure}[t]
		\begin{center}
			\includegraphics[width=1.0\linewidth]{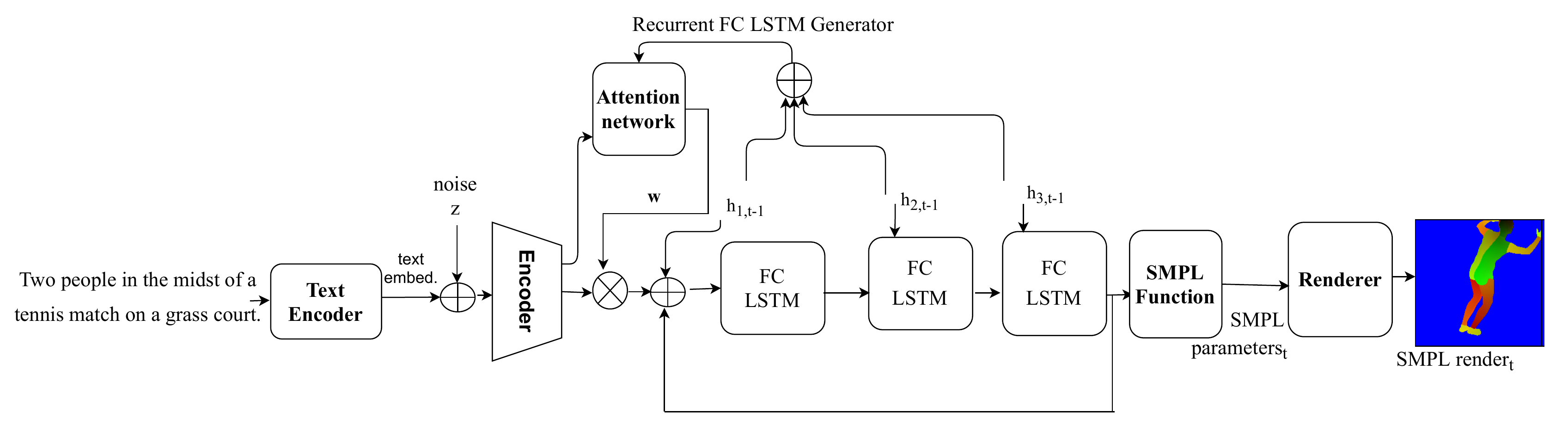}
		\end{center}
		\caption{
			First a text embedding of a caption is computed using a language model and concatenated with a latent vector $z$. An encoder network receives this input and produces multi-channel features for each word. We calculate weights for the word encodings that indicate their contribution for generating one matching person shape at iteration $t$. The weighted word features are then concatenated with the previous hidden state and the previous output.}
		\label{fig:approach_generator}
	\end{figure}
	\begin{figure}[t]
		\begin{center}
			\includegraphics[width=0.95\linewidth]{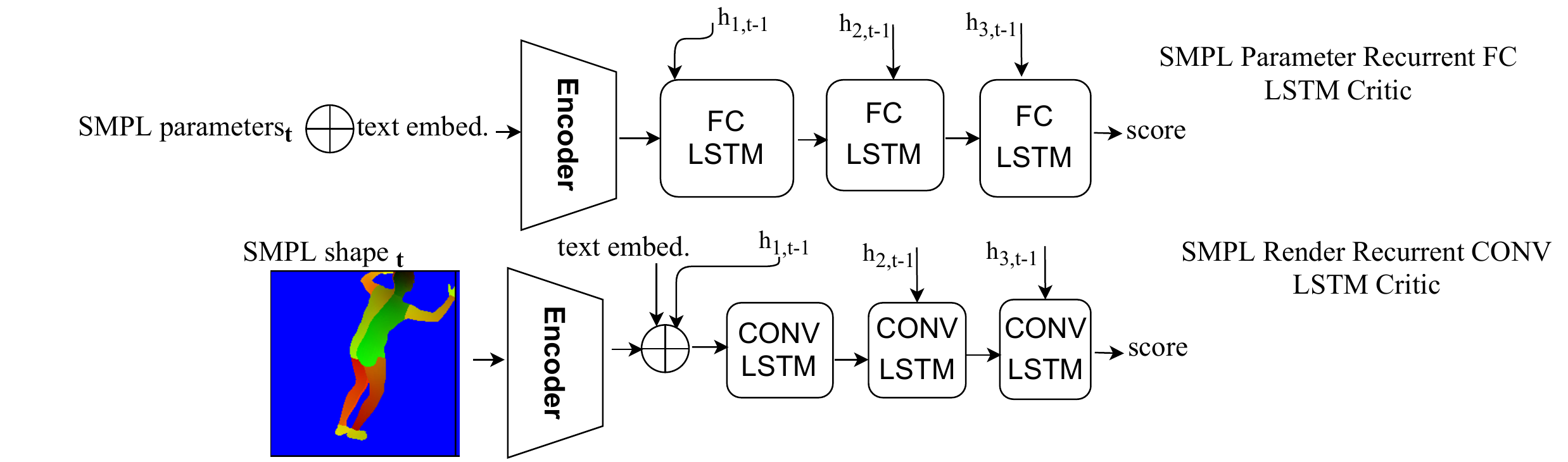}
		\end{center}
		\caption{
			The critic of the SMPL parameters takes a pair of SMPL parameters of one shape and the text embeddings in every iteration and outputs a final score after it has seen the entire set of SMPL shapes. The SMPL render critic operates in the same way, except that it takes the rendered shape of a single person which is a SMPL UV map and feeds it to an encoder network that downsamples it and concatenates it with the text embeddings before it feeds it to the LSTM part. The LSTM will output its score after having observed all the entire set.}
		\label{fig:approach_critic}
	\end{figure}
	The task of predicting a variable-sized output requires a recurrent model such as a long short-term memory (LSTM) network \cite{hochreiter1997long}. We are interested in a generative model that can generate a variable number of SMPL shapes while learning the joint probability of all individuals in a scene based on text. To achieve this, we rely on a recurrent adversarial learning scheme in which we design a recurrent iterative GAN that consists of two modules, namely the generator and the critic. In this learning paradigm, the generator and critic are two neural networks trained simultaneously with an inverse objective, in which the critic aims to differentiate between real and generated samples while the generator aims to fool the critic.
	\subsection{SMPL Body Model}
	For the 3D body model, we use the prevalent Skinned Multi-Person Linear (SMPL) model, which is a statistical human body model that allows a compact representation using shape and pose parameters. The parameters consist of pose parameters $\theta\in\R^{72}$, shape parameters $\beta\in\R^{10}$ and weak-perspective camera parameters $K\in\R^3$. The model outputs a 3D mesh using a learned differentiable function $M(\theta, \beta)$ trained on 3D scans of people, and produces vertices $V\in R^{N_v\times3}, N_v=6890$. In order to find the SMPL fittings, we use the SPIN model \cite{kolotouros2019learning} which exploits the good sides of both a deep network and an iterative optimization approach to estimate the 3D human pose and shape. SPIN uses a deep network to initialize an iterative optimization algorithm and in turn uses its output to supervise the network.
	\subsection{LSTM GAN}
	The proposed generative model is a Long Short Term Memory (LSTM) GAN that consists of a recurrent generator and two recurrent critics. Such a recurrent design is critical for solving the task, since the output is a set of SMPL shapes with a variable size. Additonally, when generating a set of SMPL shapes that are conditioned on a given text, the generator needs to have a memory component in order to monitor for which words the SMPL shapes have already been generated, and which words still need to be tended to in order to avoid repetitions. The critic is designed to be recurrent in the same way as the generator. This is because when the critic scores a given set of SMPL parameters conditioned on text, first it should be able to handle a variable-length set. Additionally, several aspects need to be assessed simultaneously, these include the realism of the shape, its semantic consistency with the text and whether all parts of the text have a matching shape. That is, when a pair of a SMPL shape set and text embedding $(s, x)$ is assessed, the critic needs a memory component in order to output a final score that takes into account all these criteria. Since for example, it is possible that most shapes match the input text but perhaps a shape is missing or a duplicate shape exists or a non-matching shape is generated. Therefore, the entire shape set needs to be seen by the critic in order to be evaluated as in a many-to-one recurrent LSTM. The input to every iteration of the critic is a concatenation of the word embeddings and the SMPL shape generated in the same iteration $t$ in the generator or a single SMPL real shape . The critics are illustrated in figure \ref{fig:approach_critic}. We use the Wasserstein GAN (WGAN) variant \cite{arjovsky2017wasserstein} which uses the earth mover's distance and makes the training more stable than the Vanilla GAN.
	\subsubsection{Critics} 
	The proposed model consists of two recurrent critics. The first critic adds a direct supervisory signal on the set of SMPL parameters $s\in R^{k\cdot 85}$ where $k$ is the set size (number of person shapes). It has an encoder network $(Enc(.))$ for encoding the input which is a concatenation of a pair of the SMPL parameters set $s$ and text embedding $x$. It additionally has a recurrent fully connected LSTM network whose input in every iteration is a the encoded pair $Enc(s_t)$ corresponding to the SMPL parameters of one person $s_t$ generated in iteration $t$, and the text embedding $x$. The LSTM is iterated $k$ times where $k$ is the number of persons appearing in the image. Given the SMPL parameters alone which are simply represented as a real-valued vector, it is hard to distinguish whether a generated shape of the SMPL parameters ultimately corresponds to a realistic human shape or not and the network cannot learn the SMPL function which is responsible for generating the 3D body representation implicitly. Accordingly, in order to simplify the generator's task, we add an additional critic whose input is the rendered SMPL UV map images. These maps are downsampled using several convolutional layers and then each map's features are separately concatenated with the text embeddings and fed to the recurrent part of the critic. The rendered map is obtained from the set of vertices $V$ using a differentiable renderer that projects them onto the $2D$ space. This makes it possible to backpropagate through this part of the network as well, thereby making the entire pipeline differentiable. The advantage of these two critics is shown in the ablative studies such that when dropping either of them, the model fails to learn.
	Based on WGAN's earth mover's distance, the critic's objective is to maximize the distance between the true and generated samples. This is reflected in the first term in eq.~\ref{eq:loss_D1}. Additionally, the critic should be able to discern when a pair of a SMPL set and text embeddings is real but is mismatched. This is reflected in the second term of eq.~\ref{eq:loss_D1}. WGAN's objective requires the critic to be Lipschitz continuous, which is guaranteed when the critic's gradient norm is constrained to $1$ \cite{gulrajani17}. The Lipschitz penalty term corresponds to the third term in eq.\ref{eq:loss_D1} and its term is expressed in eq.~\ref{eq:loss_lp}.
	
	\begin{equation}\label{eq:loss_D1}
		\begin{aligned}
			L_{D_{1}} = &-\mathbb{E}_{(s,x)\sim\mathbb{P}_r,z\sim\mathbb{P}_{z}}[D(s,x)-D(G(z,x),x)]\\
			& -\mathbb{E}_{(s,x)\sim\mathbb{P}_r,\bar{x}\sim\mathbb{P}_{x}}[D(s,x)-D(s,\bar{x})]
			\\
			& +\lambda L_{LP},
		\end{aligned}
	\end{equation}
	where $(s,x)\sim\mathbb{P}_r$ is pair of a SMPL set and the corresponding text embeddings from the training set $\mathbb{P}_r$. $G(z,x)$ is the generated SMPL set for the same text embedding $x$, and $z\sim N(0,I)$ is the latent vector. The mismatched text encoding $\bar{x}$ is randomly sampled from the training set. 
	\begin{equation}\label{eq:loss_lp}
		L_{LP}=\mathbb{E}_{(\hat{s},x)\sim\mathbb{P}_{\hat{s},x}}[(\|\nabla_{\hat{s},x}D(\hat{s},x)\|_2-1)^2],
	\end{equation}
	where $\hat{s}$ is a random interpolation between real and generated samples, a sampling which is motivated by the fact that an optical critic has gradient norm 1 at these coupled points \cite{gulrajani2017improved}. $\lambda$ is a regularization hyperparameter of the Lipschitz constraint.  
	The loss of the second critic $L_{D_2}$ is identical to the first critic loss  $L_{D_1}$, except that we replace the set of the SMPL parameters $s$ with the set of the rendered UV images $R(M(s))$, where $M(.)$ is the SMPL function and $R(.)$ is the rendering function.
	The total critic loss comprises the loss terms of both critics: $L_D=L_{D_1} + L_{D_2}$ and both terms are weighted equally.
	\subsubsection{Generator}
	As stated earlier, the textual descriptions are the condition that guides the network in generating matching SMPL parameters. In order to obtain the text features, the text descriptions are fed to a pre-trained bi-direction LSTM encoder trained with Deep Attentional modal Similarity Model (DAMSM) \cite{xu2018attngan}. This component is optimized by maximizing the similarity between an image and text features at the word level, therefore guarantees fine-grained image-text matching. 
	Since a network can only take a fixed number of channels in the input, we take a maximum of $n\in \N$ words and pad with zeros if the sentence consists of less than $n$ words, making the size of the embedding $x\in\R^{n\cdot L}$ where $L=256$ is the length of the embedding vector. Having word-level encoding provides a fine level of detail for each word separately and allows the LSTM to focus its attention on the individual word when generating the SMPL shapes. The text embeddings are concatenated with a latent vector $z\in\R^d$ that is sampled from the normal distribution. $x\oplus z$ is first fed through an encoder network that consists of several fully connected layers. 
	These features are concatenated with the LSTM hidden state of the previous time step $h_{1,t-1}$ and the previous output of the LSTM $o_{t-1}$ and fed through the LSTM part which generates the SMPL parameters  $s_t\in \R^{1\cdot85}$ for one person. The LSTM maintains a hidden state that helps with associating the generated shapes with the text selectively, as well as in monitoring for which text part a matching shape has already been generated. Both $O_{t-1}$ and  $h_{1,t-1}$ are initialized to zero in the first time step. Generally, the text description that describes a scene does not refer to a distinct number of people and several numbers could match the text (for example, several people are skiing on the mountain). Additionally, sometimes there are several people in the background which are ignored by text annotators. During training, we iterate the LSTM based on the number of annotated persons in the scene. During inference, to predict a reasonable number that can be used to halt the LSTM, we pretrain a shallow fully connected network that predicts a probability vector $p$ whose length can be set to the maximum number of annotated people in a given dataset. During inference, the number of human shapes to be generated is estimated by $\argmax_i(p)$. Alternatively, since it is not always a distinct number, it can be set to the weighted average of $\sum i\cdot p_i$.
	The recurrent part of the generator consists of a fully connected LSTM network (FC-LSTM) with several hidden layers, which regresses a set of SMPL parameters corresponding to the shape of a single person in every iteration. To aid the generator in paying attention to the words relevant to a given shape, we additionally learn attention weights $w_i$ s.t. $\sum_{i=1}^{n}w_i=1$. The weights are calculated in every LSTM iteration based on the hidden state and the encoded features of the text embeddings. The encoded embeddings are then multiplied by these weights before being fed to the recurrent part of the generator. These weights are weakly supervised using the critic's loss term. The generator loss term is given by:
	
	\begin{equation}\label{eq:loss_g_param}
		\begin{aligned}
			L_{G_{1}}=&-\mathbb{E}_{z\sim\mathbb{P}_{z},x\sim\mathbb{P}_{x}}\left[D(G(z,x),x)\right]
			\\
		\end{aligned}
	\end{equation}
	
	\begin{equation}\label{eq:loss_g_render}
		\begin{aligned}
			L_{G_{2}}=&-\mathbb{E}_{z\sim\mathbb{P}_{z},x\sim\mathbb{P}_{x}}\left[D(R(M(G(z,x))),x)\right]
			\\
		\end{aligned}
	\end{equation}
	and the total generator loss is the sum of both loss term: $L_G=L_{G_{1}}+L_{G_{2}}$
	where the first loss term refers to the generated SMPL parameter set $G(z,x)$ and the second refers to the rendering of this set. An overview of the generator is shown in figure~\ref{fig:approach_generator}.
	
		\section{Model Architecture and Training}

	\subsubsection{Generator}
	Here we provide a detailed description of the generator architecture which is described in figure~\ref{fig:approach_generator}. First, the text encoder takes the text description and produces word embeddings of length $L=256$ for each word in the sentence. The output is therefore $x\in\R^{n\cdot L}$ where $n=17$ is the maximum number of words taken. The vector  $z\in\R^{120}\sim N(0,I)$ is repeated $L$ times and concatenated with the embedding of each word $x_i$. The motivation is to maintain a separate encoding of each word so that the LSTM can select words for which it needs to generate the matching shape, and this is made easier when the different words encodings are not compounded. The encoding of $x\oplus z$ is fed to a small encoder network with two fully connected layers. At this stage, we find the attention weights $w_i$ for each of the encoded words $i\in [n]$. The weights are learned using a fully connected network consisting of 3 layers and a softmax operation for normalizing their sum to 1 for each generated shape. The input to the attention network is a concatenation of the extracted features and the previous hidden state of the LSTM. The extracted features are then multiplied by the attention weights and concatenated with the LSTM hidden state $h_{t-1}$ and the previous output $o_{t-1}$ and inputted to the LSTM network which has 3 hidden cells with gates consisting of fully connected layers as the output is a real-valued vector representing the SMPL parameters $s_t\in \R^{85}$ of one person. 
	
	\subsubsection{Critics} 
	The first critic of the SMPL parameters (figure 4 in the manuscript) takes a pair of a set of SMPL parameters $s\in R ^{k\cdot 85}$ ($k$ is the number of person shapes) and the text embeddings $x\in\R^{17\cdot 256}$. Each SMPL element $s_t$ in the set (corresponding to a person shape) is concatenated with the text embedding and the previous hidden state $h_{t-1}$ and then fed to a two-layer encoder network. The number of extracted features correspond to the set size which is the number of human shapes to be assessed by the critic. At this stage, each of the features corresponding to a shape is fed to the LSTM part and the LSTM is iterated $|s|$ times. The final score that is taken into account in the loss function is the one outputted by the LSTM in the last iteration. The LSTM consists of 3 hidden cells with fully connected gates where the hidden size is .
	
	The second critic is similar to the first critic, but since now we have 2D images of resoltion $224\times 224$ representing the rendered human shapes rather than vectors of the SMPL parameters, the images need to be downscaled from resolution $224\times 224$ to resolution $14\times 14$  before being concatenated with the text embeddings, where the text embeddings are repeated to match the set size and  the downscaled resolution of $14$. Additionally, the LSTM part consists of 3 fully convolutional gates which further downscale the features until a real-value representing the score is produced. The weight of the Lipschitz term in eq. 1 is set to $\lambda=10$. To obtain the rendered shapes, we use an implementation of a differentiable renderer \footnote[1]{https://github.com/daniilidis-group/neural\_renderer}, which is based on Neural 3D Mesh Renderer~\cite{kato2018neural}. 
	\paragraph{Dataset.} We use the COCO (Common Objects in Context)~\cite{lin14} dataset for training and evaluating the model. This dataset contains annotated images of everyday scenes and every image has five human-written text descriptions and 2D keypoints. The SMPL fittings are extracted using SPIN~\cite{kolotouros2019learning} from images that contain humans which yields about $35k$ training sample.
	\subsection{Training}
	The pseudo-ground truth of the SMPL parameters are obtained using the SPIN framework \footnote[2]{https://github.com/nkolot/SPIN}. SPIN expects a bounding box crop of each person appearing in the image. The crops are augmented using small scale and rotation transformations. These samples comprise the real pairs of SMPL shapes and text embeddings which are used to train the critics. 
	RMSprop optimizer is used with learning rate $\eta=1e^{-5}$. Based on WGAN update strategy, the critic is updated 5 times for every generator update step. After filtering out images without a valid SPIN prediction, there are in total $35k$ (image, text) sample pairs, in which the number of persons in each image varies. The model is trained for $30$ epochs.  
	
	\section{Experiments}
	
	\subsection{Qualitative evaluation}
\begin{figure}
		\centering
		\setlength\tabcolsep{5pt}
		\scalebox{1}{
			\begin{tabular}{lll}
					\multicolumn{1}{c}{\includegraphics[scale=0.25]{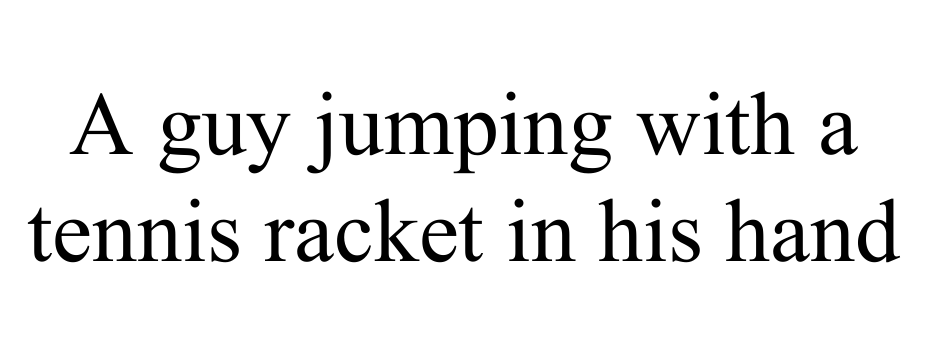}}&
					\multicolumn{1}{c}{\includegraphics[scale=0.25]{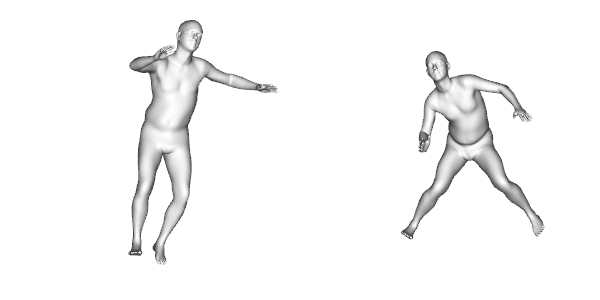}}\\  
				\multicolumn{1}{c}{\includegraphics[scale=0.25]{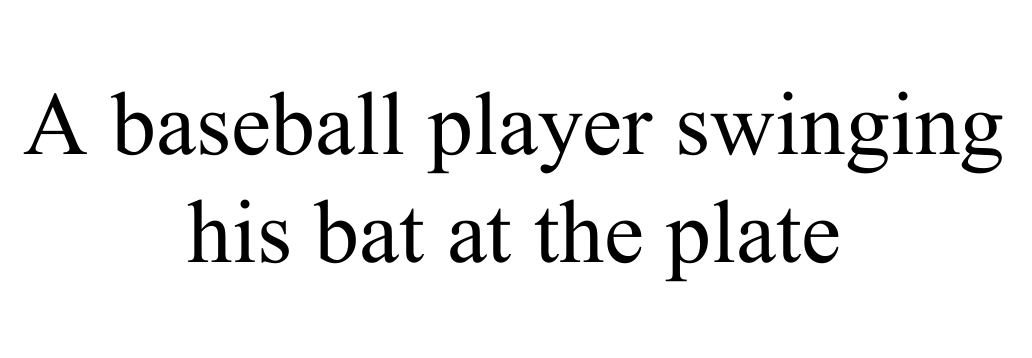}}& 
				\multicolumn{1}{c}{\includegraphics[scale=0.25]{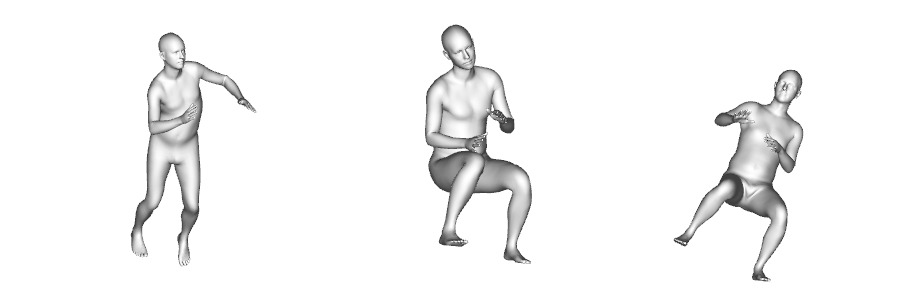}}\\  
				\multicolumn{1}{c}{\includegraphics[scale=0.25]{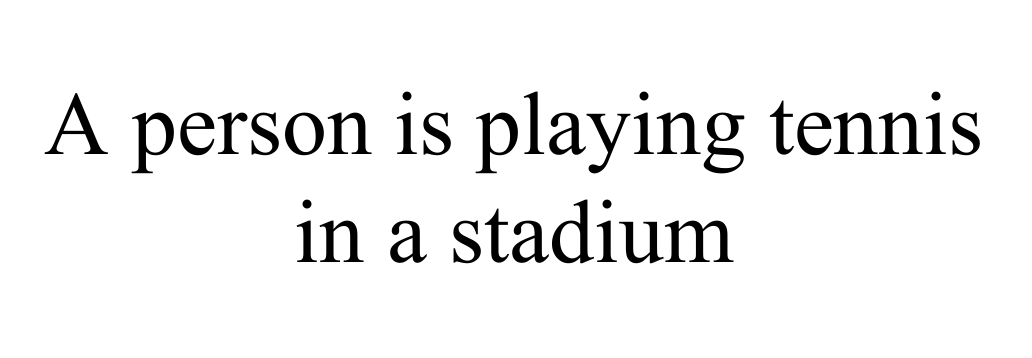}}&
				\multicolumn{1}{c}{\includegraphics[scale=0.25]{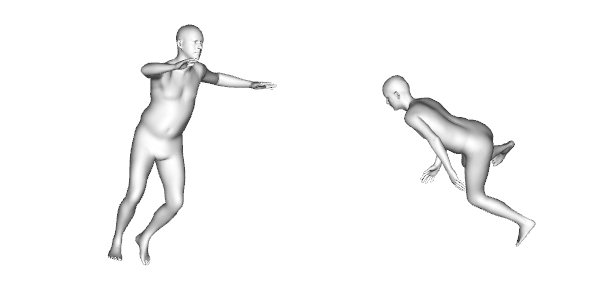}}\\  
				\multicolumn{1}{c}{\includegraphics[scale=0.25]{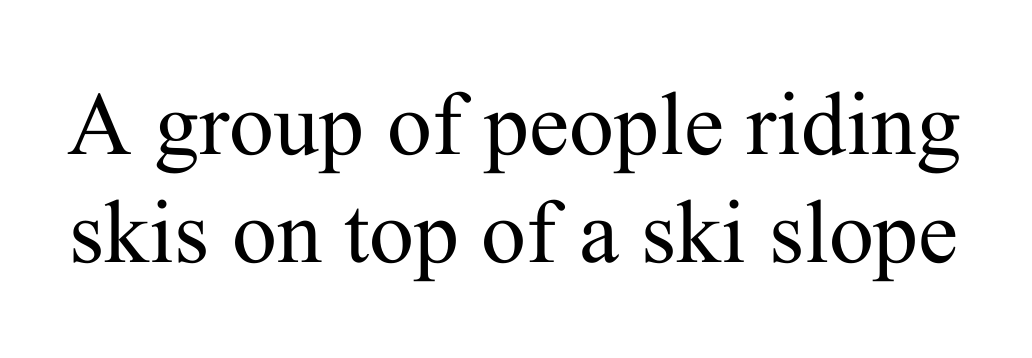}}&
				\multicolumn{1}{c}{\includegraphics[scale=0.25]{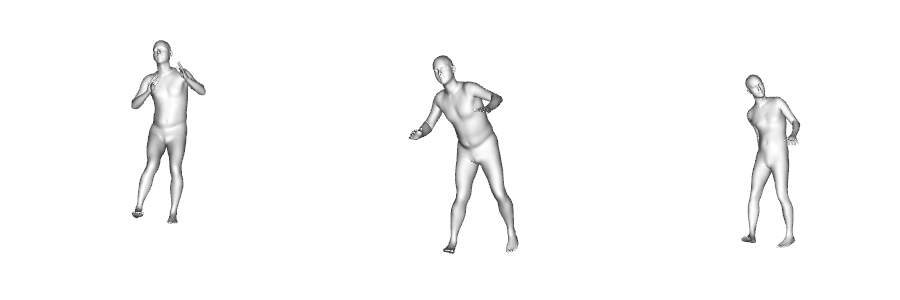}}\\  
				\multicolumn{1}{c}{\includegraphics[scale=0.25]{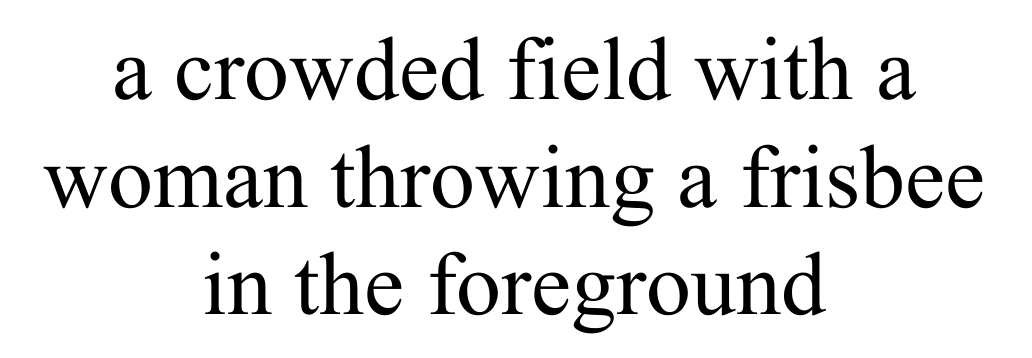}}& 
				\multicolumn{1}{c}{\includegraphics[scale=0.25]{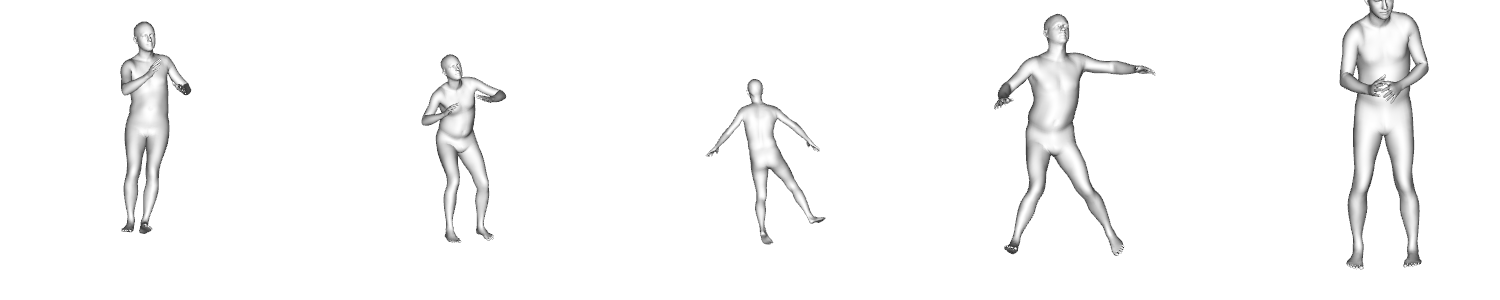}}\\      
				\multicolumn{1}{c}{\includegraphics[scale=0.25]{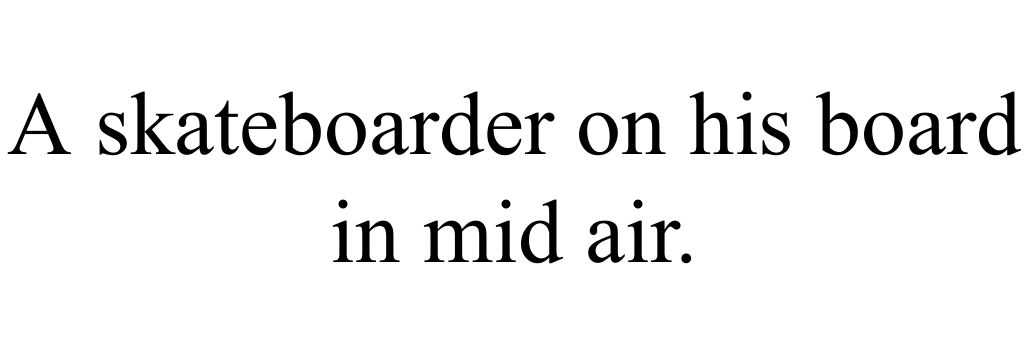}}&
				\multicolumn{1}{c}{\includegraphics[scale=0.25]{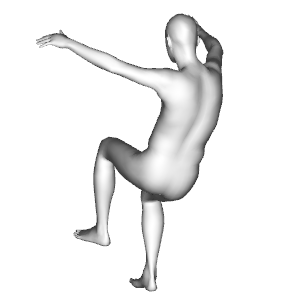}}\\ 
				\multicolumn{1}{c}{\includegraphics[scale=0.25]{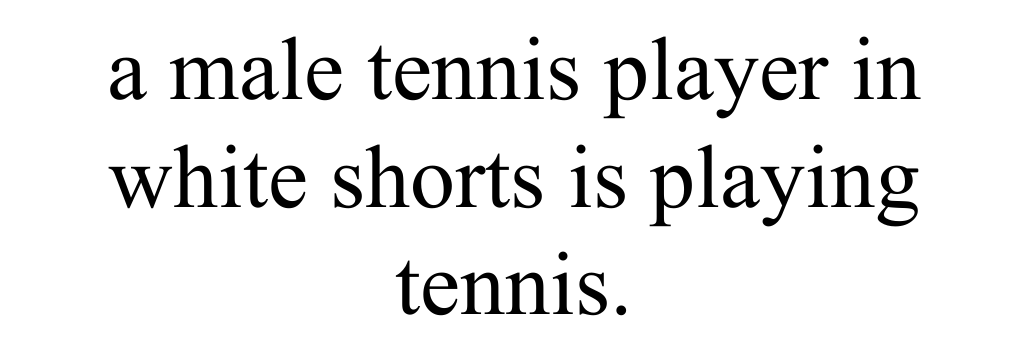}}&
				\multicolumn{1}{c}{\includegraphics[scale=0.25]{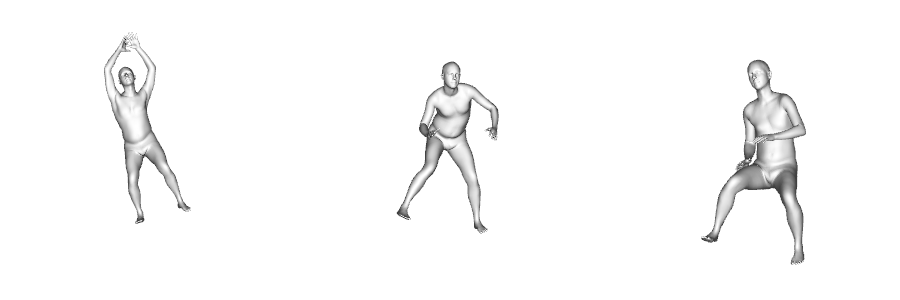}}\\
				
				\multicolumn{1}{c}{Caption}&\multicolumn{1}{c}{generated mesh}
		\end{tabular}}
		\caption{Example generated SMPL shapes from the COCO validation set. In the second row, even though the text description refers to a person jumping with a racket, the model infers an additional player since in COCO most images with a tennis match have two annotated players. The model also infers the interaction between the players, such that the player on the left is striking the ball and the player on the right is preparing to return it.}
		\label{fig:smpl_syn_examples}
	\end{figure}

	Figure \ref{fig:smpl_syn_examples} includes synthesized SMPL images. As can be seen, the synthesized SMPL shapes match the text description. 
	
\begin{figure}[t]
		\centering
		\setlength\tabcolsep{5pt}
		\scalebox{1}{
			\begin{tabular}{llll}
				\multicolumn{1}{c}{\includegraphics[scale=0.25]{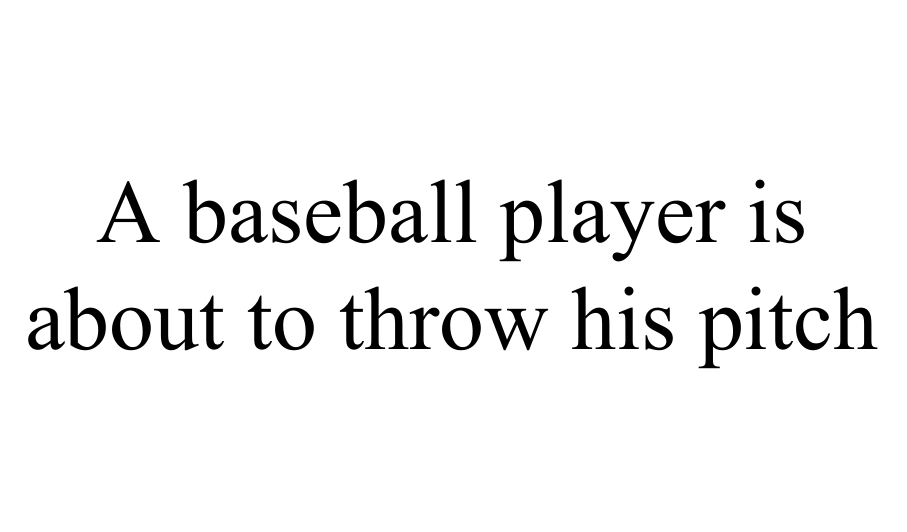}}&  
				\multicolumn{1}{c}{\includegraphics[scale=0.3]{images/image_synthesis/gt/COCO_val2014_000000026512}}&  
				\multicolumn{1}{c}{\includegraphics[scale=0.3]{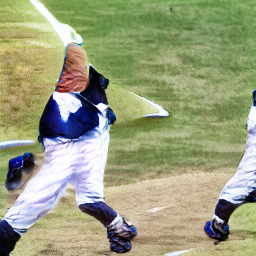}}&  
				\multicolumn{1}{c}{\includegraphics[scale=0.3]{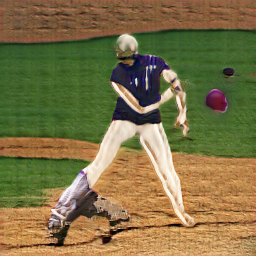}}\\ 
				\multicolumn{1}{c}{\includegraphics[scale=0.25]{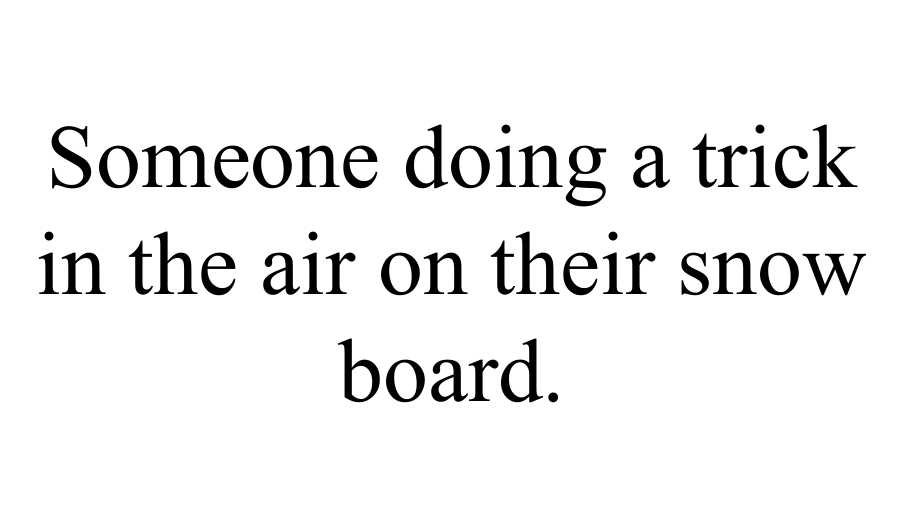}}& 
				\multicolumn{1}{c}{\includegraphics[scale=0.3]{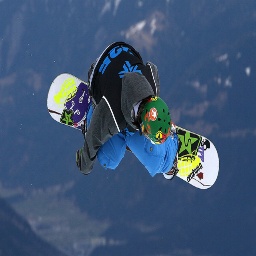}}&  
				\multicolumn{1}{c}{\includegraphics[scale=0.3]{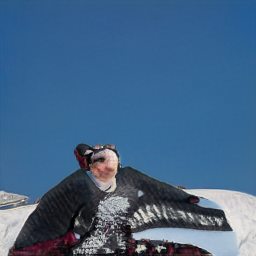}}&  
				\multicolumn{1}{c}{\includegraphics[scale=0.25]{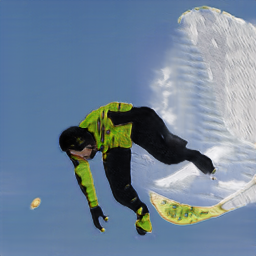}}\\
				\multicolumn{1}{c}{\includegraphics[scale=0.3]{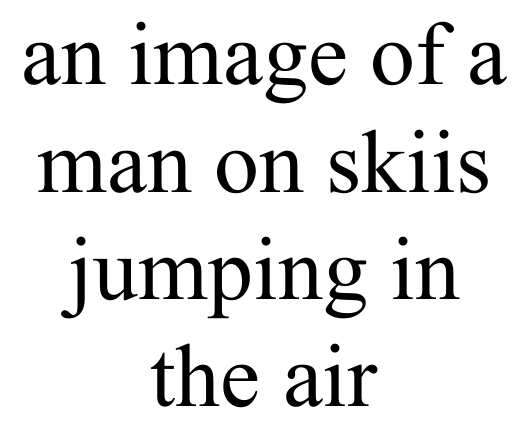}}&  
				\multicolumn{1}{c}{\includegraphics[scale=0.3]{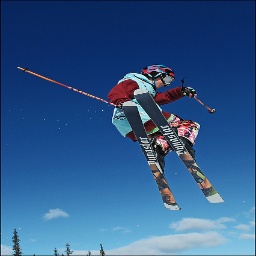}}&  
				\multicolumn{1}{c}{\includegraphics[scale=0.3]{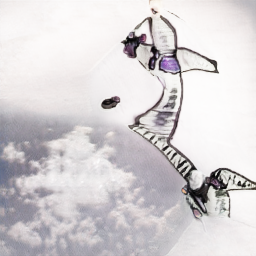}}&  
				\multicolumn{1}{c}{\includegraphics[scale=0.3]{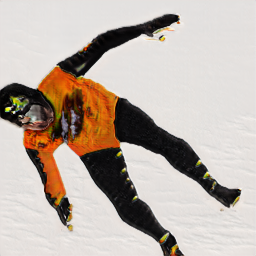}}\\ 
				\multicolumn{1}{c}{\includegraphics[scale=0.25]{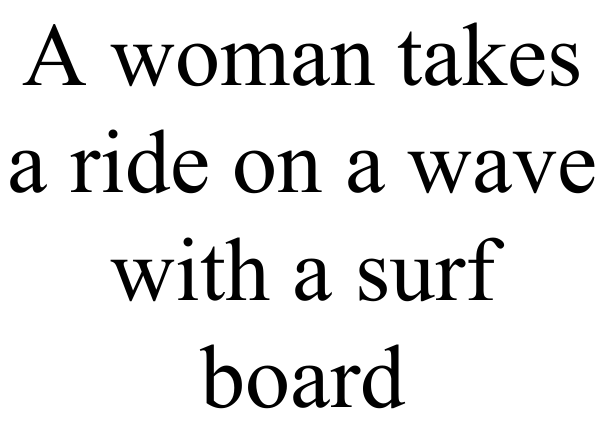}}& 
				\multicolumn{1}{c}{\includegraphics[scale=0.3]{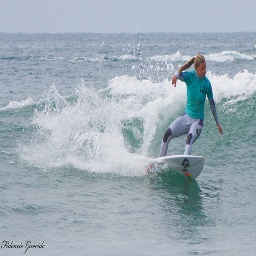}}&  
				\multicolumn{1}{c}{\includegraphics[scale=0.3]{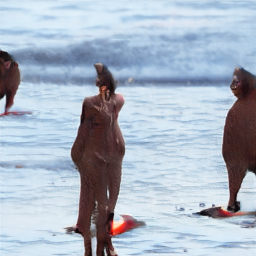}}&  
				\multicolumn{1}{c}{\includegraphics[scale=0.3]{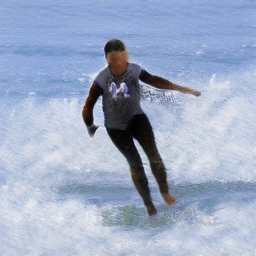}}\\
				\multicolumn{1}{c}{\includegraphics[scale=0.25]{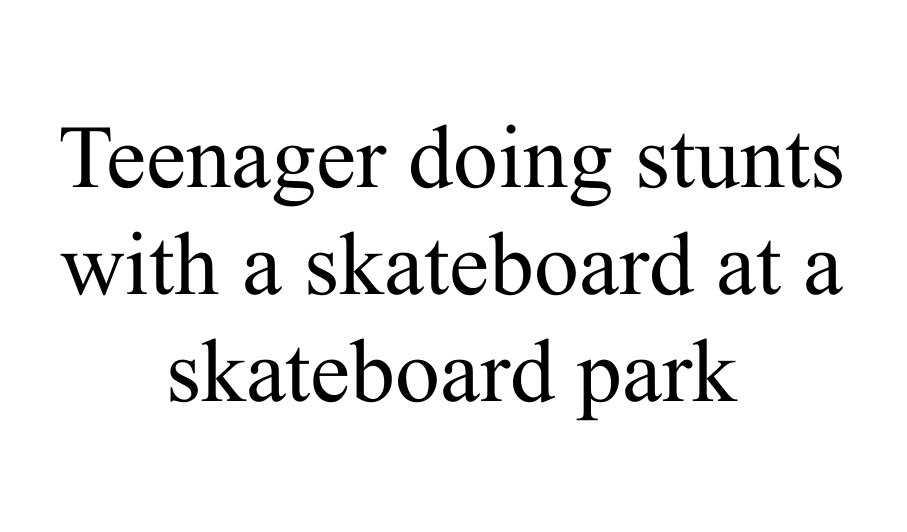}}&  
				\multicolumn{1}{c}{\includegraphics[scale=0.3]{images/image_synthesis/gt/COCO_val2014_000000581357}}&  
				\multicolumn{1}{c}{\includegraphics[scale=0.3]{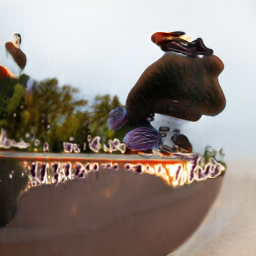}}&  
				\multicolumn{1}{c}{\includegraphics[scale=0.3]{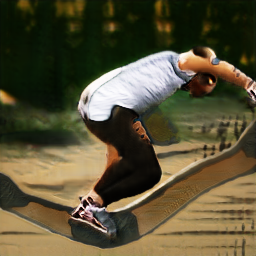}}\\ 
				\multicolumn{1}{c}{Caption} & {Ground truth}&  \multicolumn{1}{c}{DF GAN}&  \multicolumn{1}{c}{DF with SMPL}
		\end{tabular}}
		\caption{Images generated from several captions. The images generated with DF GAN \cite{tao2020df} synthesize the scene well but fail to synthesize realistic looking humans. When the generated SMPL image is injected along with the text features, it guides the network towards generating a realistic-looking person that reflects the relevant text description. It can be observed that the network further learns to create body parts such as the hand at a fine level of detail, and can distinguish the different clothing parts and places the person reasonably within the scene.}
		\label{fig:im_syn_examples}
	\end{figure}

	To show the benefit of the proposed task, we integrate our model with an image synthesis framework while using COCO as the training dataset. We injected the generated SMPL shape during the training of~\cite{tao2020df}, which is a state of the art image synthesis framework. We then evaluated on captions from the COCO validation set. Figure~\ref{fig:im_syn_examples} shows the results. We can see how first generating the human shape using our pretrained model substantially improves the person's appearance in the synthesized images.
	\subsection{Quantitative evaluation} 
	To evaluate our model quantitatively, we use measures that are similar to those introduced in \cite{zhang2021adversarial} for single pose synthesis. However, since we generate multiple shapes, we need to adapt the calculation such that shapes from two different samples are matched together. In order to find an optimal matching between two samples of a generated and real SMPL set, we use the Hungarian algorithm where the cost matrix is calculated between each pair of shapes. We sample $400$ samples randomly from the validation set to make the computation time reasonable. $\bar{d}^p_{nn}$ denotes the average distance between the generated sample and nearest neighbor real sample. $\bar{d}^t_{p_{nn}}$ denotes the distance in the embedding space of the nearest neighbor, indicating that even if the nearest neighbor sample is not the ground-truth itself, the text distance to the nearest neighbor is still small. This values is compared with $\bar{d}^t_{all}$ which denotes the average unconditional distance between all text embeddings (this distance is fixed). $\bar{d}^p_{gt}$ denotes the average distance between the generated sample and the ground-truth real sample (the one corresponding to the input text). We compute the distances in both the SMPL parameter space and UV space which are denoted by the superscript $p/r$ and the results are presented in tables~\ref{tab:quantitaive_eval_param},~\ref{tab:quantitaive_eval_uv} respectively. 
	
	\begin{table*}
		\begin{center}
			\begin{tabular}[h]{l|cc|cc}
				\hline
				&parameter distance & &&text distance\\
				&$\bar{d}^p_{nn}$ & $\bar{d}^p_{gt}$ & $\bar{d}^t_{p_{nn}}$ 
				&$\bar{d}^t_{all}$ \\  
				\hline
				both critics&$1.66$ & $3.59$ &$6.82$&$6.92$\\
				wo critic2&$1.78$ & $3.72$ &$6.91$&$6.92$

			\end{tabular}
		\end{center}
		\caption{Quantitative evaluation of the generated SMPL shapes in the SMPL parameter space. `wo critic2` refers to dropping the critic of the rendered SMPL shape.
		} 
		\label{tab:quantitaive_eval_param}
	\end{table*}

	\begin{table*}
	\begin{center}
		\begin{tabular}[h]{l|cc|cc}
			\hline
			&UV distance & &&text distance\\
			&$\bar{d}^r_{nn}$ & $\bar{d}^r_{gt}$ & $\bar{d}^t_{p_{nn}}$ 
			&$\bar{d}^t_{all}$ \\  
			\hline
			both critics&$51.89$ & $72.19$  &$6.72$&$6.92$\\
			wo critic2&$55.23$ & $74.5$ &$6.91$&$6.92$\\

			
		\end{tabular}
	\end{center}
	\caption{Quantitative evaluation of the generated SMPL shapes in the SMPL UV space. "wo critic2" refers to dropping the critic of the rendered SMPL shape.
	} 
	\label{tab:quantitaive_eval_uv}
\end{table*}
	\subsection{User study}
	In order to evaluate the realism of the generated shapes, we created a user study with $10$ questions in which we display a set of generated shapes and $3$ different captions that include one ground truth caption and another two captions. Users were asked to select the caption that best matches the SMPL render set. The average correct score was $68\%$. Sometimes it is hard to determine distinctly what text the shape corresponds to if the action is generic (such as standing), unlike shapes for distinct activities such as batting a ball or skating. However, such details can be determined more confidently once the context is clear from the whole scene as shown in the synthesized images in figure~\ref{fig:im_syn_examples}.
	
	\subsection{Ablative studies}
	If the SMPL parameters critic is dropped, then we lose the direct supervisory signal on the GAN output and the model fails to learn completely. When the SMPL render critic is dropped, then we lose the supervisory signal on the human shape and the rendered shapes quality degrades. This is also reflected in the quantitative measures in table~\ref{tab:quantitaive_eval_param} and~\ref{tab:quantitaive_eval_uv}.
	
	When perturbing the word ordering such as in an inverse or random ordering, then the model diverges completely, indicating that the word ordering is crucial for learning the relationship between the generated SMPL shapes and the text.

	\subsection{Conclusion}
	In this work we have introduced the task of generating multiple 3D shapes of the human body conditioned on text. We have evaluated the models using different metrics and a user study. We additionally have demonstrated how it helps   synthesizing more realistic and refined human shapes in image synthesis frameworks.

	\paragraph{Acknowledgement}
	We would like to thank Dimitris Tzionas for providing insightful and valuable feedback on this work.
	
	\clearpage
	\bibliographystyle{splncs04}
	\bibliography{egbib}

\begin{thebibliography}{10}
\providecommand{\url}[1]{\texttt{#1}}
\providecommand{\urlprefix}{URL }
\providecommand{\doi}[1]{https://doi.org/#1}

\bibitem{arjovsky2017wasserstein}
Arjovsky, M., Chintala, S., Bottou, L.: Wasserstein generative adversarial
  networks. In: International conference on machine learning. pp. 214--223.
  PMLR (2017)

\bibitem{chu2017cyclegan}
Chu, C., Zhmoginov, A., Sandler, M.: Cyclegan, a master of steganography. arXiv
  preprint arXiv:1712.02950  (2017)

\bibitem{dai2017towards}
Dai, B., Fidler, S., Urtasun, R., Lin, D.: Towards diverse and natural image
  descriptions via a conditional gan. In: Proceedings of the IEEE International
  Conference on Computer Vision. pp. 2970--2979 (2017)

\bibitem{esteban2017real}
Esteban, C., Hyland, S.L., R{\"a}tsch, G.: Real-valued (medical) time series
  generation with recurrent conditional gans. arXiv preprint arXiv:1706.02633
  (2017)

\bibitem{goodfellow14}
Goodfellow, I., Pouget-Abadie, J., Mirza, M., Xu, B., Warde-Farley, D., Ozair,
  S., Courville, A., Bengio, Y.: Generative adversarial nets. Advances in
  Neural Information Processing Systems  \textbf{27} (2014)

\bibitem{grabner2011makes}
Grabner, H., Gall, J., Van~Gool, L.: What makes a chair a chair? In: CVPR 2011.
  pp. 1529--1536. IEEE (2011)

\bibitem{gulrajani2017improved}
Gulrajani, I., Ahmed, F., Arjovsky, M., Dumoulin, V., Courville, A.: Improved
  training of wasserstein gans. arXiv preprint arXiv:1704.00028  (2017)

\bibitem{gulrajani17}
Gulrajani, I., Ahmed, F., Arjovsky, M., Dumoulin, V., Courville, A.C.: Improved
  training of wasserstein gans. In: Advances in Neural Information Processing
  Systems (2017)

\bibitem{gupta20113d}
Gupta, A., Satkin, S., Efros, A.A., Hebert, M.: From 3d scene geometry to human
  workspace. In: CVPR 2011. pp. 1961--1968. IEEE (2011)

\bibitem{hinton2009deep}
Hinton, G.E.: Deep belief networks. Scholarpedia  \textbf{4}(5), ~5947 (2009)

\bibitem{hochreiter1997long}
Hochreiter, S., Schmidhuber, J.: Long short-term memory. Neural computation
  \textbf{9}(8),  1735--1780 (1997)

\bibitem{kato2018neural}
Kato, H., Ushiku, Y., Harada, T.: Neural 3d mesh renderer. In: Proceedings of
  the IEEE conference on computer vision and pattern recognition. pp.
  3907--3916 (2018)

\bibitem{kingma2013auto}
Kingma, D.P., Welling, M.: Auto-encoding variational bayes. arXiv preprint
  arXiv:1312.6114  (2013)

\bibitem{kolotouros2019learning}
Kolotouros, N., Pavlakos, G., Black, M.J., Daniilidis, K.: Learning to
  reconstruct 3d human pose and shape via model-fitting in the loop. In:
  Proceedings of the IEEE/CVF International Conference on Computer Vision. pp.
  2252--2261 (2019)

\bibitem{li2019object}
Li, W., Zhang, P., Zhang, L., Huang, Q., He, X., Lyu, S., Gao, J.:
  Object-driven text-to-image synthesis via adversarial training. In:
  Proceedings of the IEEE Conference on Computer Vision and Pattern
  Recognition. pp. 12174--12182 (2019)

\bibitem{Li_2019_affordance}
Li, X., Liu, S., Kim, K., Wang, X., Yang, M.H., Kautz, J.: Putting humans in a
  scene: Learning affordance in 3d indoor environments. In: The IEEE Conference
  on Computer Vision and Pattern Recognition (CVPR) (June 2019)

\bibitem{Li_2019_story}
Li, Y., Gan, Z., Shen, Y., Liu, J., Cheng, Y., Wu, Y., Carin, L., Carlson, D.,
  Gao, J.: Storygan: A sequential conditional gan for story visualization. In:
  The IEEE Conference on Computer Vision and Pattern Recognition (CVPR) (June
  2019)

\bibitem{lin2014microsoft}
Lin, T.Y., Maire, M., Belongie, S., Hays, J., Perona, P., Ramanan, D.,
  Doll{\'a}r, P., Zitnick, C.L.: Microsoft coco: Common objects in context. In:
  European conference on computer vision. pp. 740--755. Springer (2014)

\bibitem{lin14}
Lin, T.Y., Maire, M., Belongie, S., Hays, J., Perona, P., Ramanan, D., Dollar,
  P., Zitnick, C.L.: Microsoft coco: common objects in context. In: European
  Conference on Computer Vision (2014)

\bibitem{loper2015smpl}
Loper, M., Mahmood, N., Romero, J., Pons-Moll, G., Black, M.J.: Smpl: A skinned
  multi-person linear model. ACM transactions on graphics (TOG)
  \textbf{34}(6),  1--16 (2015)

\bibitem{ma2017pose}
Ma, L., Jia, X., Sun, Q., Schiele, B., Tuytelaars, T., Van~Gool, L.: Pose
  guided person image generation. arXiv preprint arXiv:1705.09368  (2017)

\bibitem{ma17}
Ma, L., Jia, X., Sun, Q., Schiele, B., Tuytelaars, T., Van~Gool, L.: Pose
  guided person image generation. In: Advances in Neural Information Processing
  Systems (2017)

\bibitem{mogren2016c}
Mogren, O.: C-rnn-gan: Continuous recurrent neural networks with adversarial
  training. arXiv preprint arXiv:1611.09904  (2016)

\bibitem{prokudin2021smplpix}
Prokudin, S., Black, M.J., Romero, J.: Smplpix: Neural avatars from 3d human
  models. In: Proceedings of the IEEE/CVF Winter Conference on Applications of
  Computer Vision. pp. 1810--1819 (2021)

\bibitem{NIPS2019_8375}
Qiao, T., Zhang, J., Xu, D., Tao, D.: Learn, imagine and create: Text-to-image
  generation from prior knowledge. In: Wallach, H., Larochelle, H.,
  Beygelzimer, A., d~Alche-Buc, F., Fox, E., Garnett, R. (eds.) Advances in
  Neural Information Processing Systems 32, pp. 887--897. Curran Associates,
  Inc. (2019)

\bibitem{qiao2019mirrorgan}
Qiao, T., Zhang, J., Xu, D., Tao, D.: Mirrorgan: Learning text-to-image
  generation by redescription. In: Proceedings of the IEEE Conference on
  Computer Vision and Pattern Recognition. pp. 1505--1514 (2019)

\bibitem{radford16}
Radford, A., Metz, L., Chintala, S.: Unsupervised representation learning with
  deep convolutional generative adversarial networks. In: International
  Conference on Learning Representations (2016)

\bibitem{ramesh2021zero}
Ramesh, A., Pavlov, M., Goh, G., Gray, S., Voss, C., Radford, A., Chen, M.,
  Sutskever, I.: Zero-shot text-to-image generation. arXiv preprint
  arXiv:2102.12092  (2021)

\bibitem{reed16}
Reed, S., Akata, Z., Yan, X., Logeswaran, L., Schiele, B., Lee, H.: Generative
  adversarial text to image synthesis. In: International Conference on Machine
  Learning (2016)

\bibitem{NIPS2016_a8f15eda}
Reed, S.E., Akata, Z., Mohan, S., Tenka, S., Schiele, B., Lee, H.: Learning
  what and where to draw. In: Lee, D., Sugiyama, M., Luxburg, U., Guyon, I.,
  Garnett, R. (eds.) Advances in Neural Information Processing Systems.
  vol.~29. Curran Associates, Inc. (2016),
  \url{https://proceedings.neurips.cc/paper/2016/file/a8f15eda80c50adb0e71943adc8015cf-Paper.pdf}

\bibitem{NIPS2016_6111}
Reed, S.E., Akata, Z., Mohan, S., Tenka, S., Schiele, B., Lee, H.: Learning
  what and where to draw. In: Lee, D.D., Sugiyama, M., Luxburg, U.V., Guyon,
  I., Garnett, R. (eds.) Advances in Neural Information Processing Systems 29,
  pp. 217--225. Curran Associates, Inc. (2016),
  \url{http://papers.nips.cc/paper/6111-learning-what-and-where-to-draw.pdf}

\bibitem{salakhutdinov2015learning}
Salakhutdinov, R.: Learning deep generative models. Annual Review of Statistics
  and Its Application  \textbf{2},  361--385 (2015)

\bibitem{salakhutdinov2009deep}
Salakhutdinov, R., Hinton, G.: Deep boltzmann machines. In: Artificial
  intelligence and statistics. pp. 448--455 (2009)

\bibitem{tan2019semantics}
Tan, H., Liu, X., Li, X., Zhang, Y., Yin, B.: Semantics-enhanced adversarial
  nets for text-to-image synthesis. In: Proceedings of the IEEE/CVF
  International Conference on Computer Vision. pp. 10501--10510 (2019)

\bibitem{tang2020bipartite}
Tang, H., Bai, S., Torr, P.H., Sebe, N.: Bipartite graph reasoning gans for
  person image generation. arXiv preprint arXiv:2008.04381  (2020)

\bibitem{tang2020xinggan}
Tang, H., Bai, S., Zhang, L., Torr, P.H., Sebe, N.: Xinggan for person image
  generation. In: European Conference on Computer Vision. pp. 717--734.
  Springer (2020)

\bibitem{tao2020df}
Tao, M., Tang, H., Wu, S., Sebe, N., Jing, X.Y., Wu, F., Bao, B.: Df-gan: Deep
  fusion generative adversarial networks for text-to-image synthesis. arXiv
  preprint arXiv:2008.05865  (2020)

\bibitem{xu2018attngan}
Xu, T., Zhang, P., Huang, Q., Zhang, H., Gan, Z., Huang, X., He, X.: Attngan:
  Fine-grained text to image generation with attentional generative adversarial
  networks. In: Proceedings of the IEEE conference on computer vision and
  pattern recognition. pp. 1316--1324 (2018)

\bibitem{zhang17}
Zhang, H., Xu, T., Li, H., Zhang, S., Wang, X., Huang, X., Metaxas, D.N.:
  Stackgan: Text to photo-realistic image synthesis with stacked generative
  adversarial networks. In: IEEE International Conference on Computer Vision
  (2017)

\bibitem{zhang2020generating}
Zhang, Y., Hassan, M., Neumann, H., Black, M.J., Tang, S.: Generating 3d people
  in scenes without people. In: Proceedings of the IEEE/CVF Conference on
  Computer Vision and Pattern Recognition. pp. 6194--6204 (2020)

\bibitem{zhang2021adversarial}
Zhang, Y., Briq, R., Tanke, J., Gall, J.: Adversarial synthesis of human pose
  from text. In: Pattern Recognition: 42nd DAGM German Conference, DAGM GCPR
  2020, T{\"u}bingen, Germany, September 28--October 1, 2020, Proceedings 42.
  pp. 145--158. Springer (2021)

\bibitem{zhou2019text}
Zhou, X., Huang, S., Li, B., Li, Y., Li, J., Zhang, Z.: Text guided person
  image synthesis. In: Proceedings of the IEEE Conference on Computer Vision
  and Pattern Recognition. pp. 3663--3672 (2019)

\bibitem{zhu2019dm}
Zhu, M., Pan, P., Chen, W., Yang, Y.: Dm-gan: Dynamic memory generative
  adversarial networks for text-to-image synthesis. In: Proceedings of the
  IEEE/CVF Conference on Computer Vision and Pattern Recognition. pp.
  5802--5810 (2019)

\bibitem{Zhu_2017_ICCV}
Zhu, S., Urtasun, R., Fidler, S., Lin, D., Change~Loy, C.: Be your own prada:
  Fashion synthesis with structural coherence. In: Proceedings of the IEEE
  International Conference on Computer Vision (ICCV) (Oct 2017)

\end{thebibliography}
	
\end{document}